\def\year{2020}\relax
\documentclass[letterpaper]{article} 
\usepackage{aaai20}  
\usepackage{times}  
\usepackage{helvet} 
\usepackage{courier} 
\usepackage{threeparttable}
\usepackage [usenames, dvipsnames] {color}
\usepackage[hyphens]{url} 
\usepackage{graphicx} 
\usepackage[switch]{lineno}
\usepackage{graphicx}  
\usepackage{color}

\title{Video Cloze Procedure for Self-Supervised Spatio-Temporal Learning}
\author{
Dezhao Luo,\textsuperscript{\rm 1,\rm 2}\thanks{Equal contribution}
Chang Liu,\textsuperscript{\rm 3}\footnotemark[1] 
Yu Zhou,\textsuperscript{\rm 1}\thanks{Corresponding authors} 
Dongbao Yang,\textsuperscript{\rm 1} 
Can Ma,\textsuperscript{\rm 1} 
Qixiang Ye,\textsuperscript{\rm 3}\footnotemark[2] 
Weiping Wang\textsuperscript{\rm 1}\\\\
\textsuperscript{1}{Institute of Information Engineering, Chinese Academy of Sciences}\\
\textsuperscript{2}{School of Cyber Security, University of Chinese Academy of Sciences}\\
\textsuperscript{3}{University of Chinese Academy of Sciences}\\
\{luodezhao,zhouyu,yangdongbao,macan,wangweiping\}@iie.ac.cn\\
 liuchang615@mails.ucas.ac.cn, qxye@ucas.ac.cn
}

\begin{document}

\maketitle

\begin{abstract}

We propose a novel self-supervised method, referred to as Video Cloze Procedure (VCP), to learn rich spatial-temporal representations. 
VCP first generates ``blanks" by withholding video clips and then creates ``options" by applying spatio-temporal operations on the withheld clips. Finally, it fills the blanks with ``options" and learns representations by predicting the categories of operations applied on the clips. 
%
VCP can act as either a proxy task or a target task in self-supervised learning. 
As a proxy task, it converts rich self-supervised representations into video clip operations (options), which enhances the flexibility and reduces the complexity of representation learning.
As a target task, it can assess learned representation models in a uniform and interpretable manner.
With VCP, we train spatial-temporal representation models (3D-CNNs) and apply such models on action recognition and video retrieval tasks.
Experiments on commonly used benchmarks show that the trained models outperform the state-of-the-art self-supervised models with significant margins.  
\end{abstract}

\begin{figure}[!t]
\centering
\includegraphics[width=0.9\columnwidth]{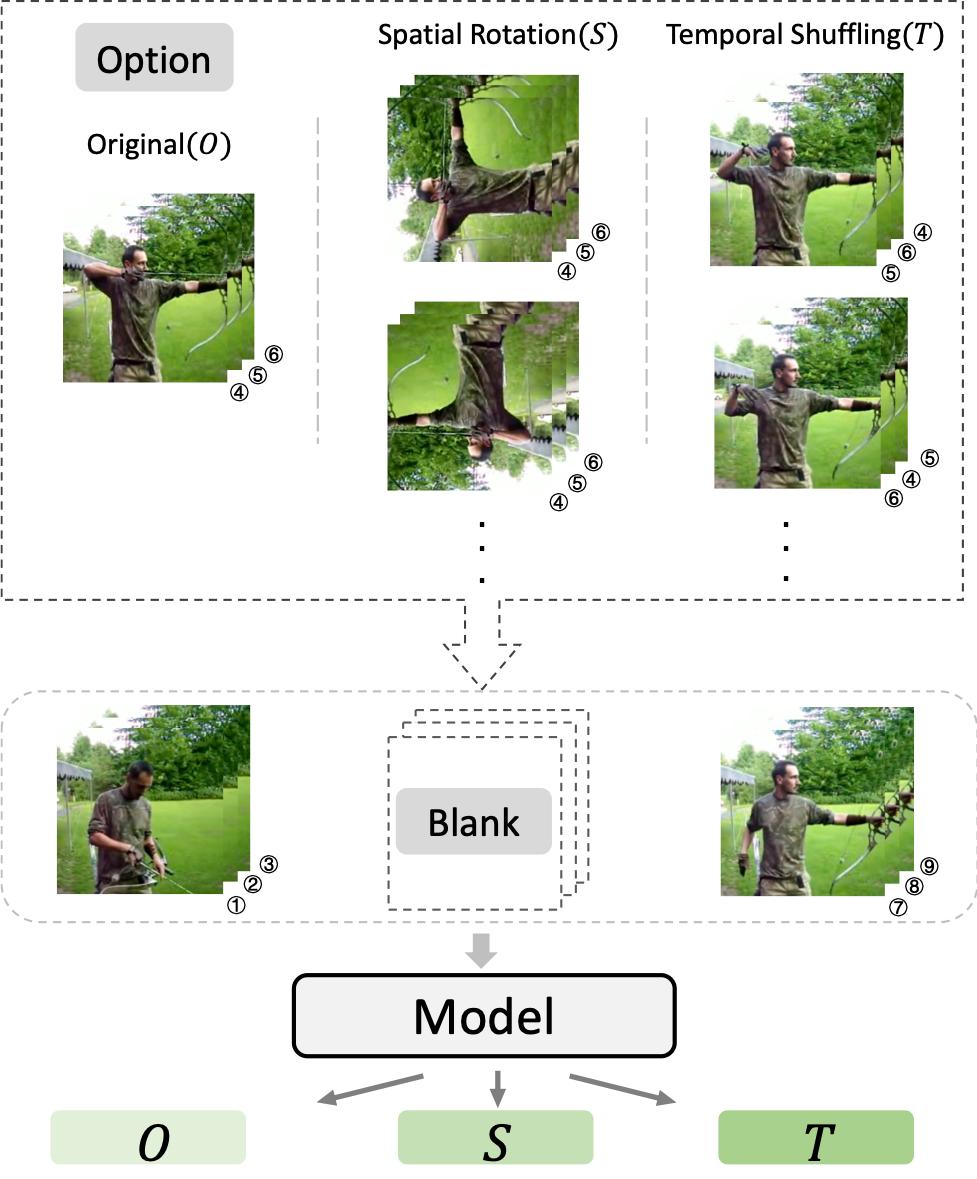}
\caption{VCP is a novel self-supervised method for spatial-temporal representation learning. It generates ``blanks" by withholding video clips, creates ``options" by applying spatial-temporal operations on the withheld clips, and completes cloze for feature learning. }
\label{fig:vcp}
\end{figure}

\section{Introduction}

In the past few years, Convolutional Neural Networks (CNNs) have unprecedentedly advanced the field of computer vision. 
Generally, vision tasks are solved by training models on large-scale datasets with label annotations~\cite{kim2019self}. Typically, CNNs pre-trained on ImageNet~\cite{ImageNet16} incorporate rich representation capability and have been widely used as initial models.

Nevertheless, annotating large-scale datasets is costly and labor-intensive, particularly when facing tasks involving complex data ($e.g.$, videos) and concepts ($e.g.$, action analysis and video retrieval)~\cite{fernando2017self,kay2017kinetics}.

To conquer this issue, self-supervised representation learning, which leverages the information from unlabelled data to train desired models, has attracted increasing attention from the artificial intelligence community.  
For video data, existing approaches usually define an annotation-free proxy task, which provides special supervision for model learning by fulfilling the objective of the proxy task. 

In the early research~\cite{doersch2015unsupervised,WangXiaoLong2015}, relative location of the patches in images or the order of video frames were used as a supervisory signal. However, the learned features were merely on a frame-by-frame basis, which are implausible to video analytic tasks where spatio-temporal features are prevailing. Recently, ~\cite{wang2019self} proposed to learn representations by regressing motion and appearance statistics. In~\cite{fernando2017self}, 
an odd-one-out network is proposed to identify the unrelated or odd video clips from a set of otherwise related clips. To find the odd video clip, the models have to learn spatio-temporal features that can discriminate video clips of minor differences.%

Despite of the effectiveness, existing approaches are usually developed upon domain-knowledge and therefore are not capable to incorporate various spatial-temporal operations. This seriously restricts the representation capability of learned models. Furthermore, the lack of a model assessment approach strikingly limits the pertinence of self-supervised representation learning.

In this paper, we propose a new self-supervised method called Video Cloze Procedure (VCP). In VCP, we withhold a video clip from a video sequence and apply multiple spatio-temporal operations on it. We train a 3D-CNN model to identify the category of operations, which drives learning rich feature representations. The motivation behinds VCP lies in that applying richer operations on video clips facilities exploring higher representation capability, Fig.\ \ref{fig:vcp}.

VCP consists of three components including blank generation, option creation, and cloze completion. The first component generates blanks by withholding video clips from given clip sequences. The second component facilitates multiple spatial-temporal representation learning by applying spatial-temporal operations on the withheld clips. Finally, cloze completion fills the blanks with options and learns representations by predicting the category of operations. 

VCP can act as either a proxy task or a target task in self-supervised learning. As a proxy task, it converts rich self-supervised representations into video clip operations, which enhances the flexibility and reduces the complexity of representation learning. As a target task, it can assess learned representation models in an interpretable manner.

The contributions of this work are summarized as follows:
\begin{itemize}

    \item We propose Video Cloze Procedure (VCP), providing a simple-yet-effective framework for self-supervised spatio-temporal representation learning.

    \item We propose a new model assessment approach by designing VCP as a special target task, which improves the interpretability of self-supervised representation learning. 
    \item VCP is applied on three kinds of 3D CNN models and two target tasks including action recognition and video retrieval, and improves the state-of-the-arts with significant margins.
\end{itemize}

\section{Related work}
Self-supervised learning leverages the information from unlabelled data to train target models. Existing approaches usually define an annotation-free proxy task which demands a network predicting information latent in unannotated videos.
The learned models can be applied to target tasks (supervised or unsupervised) in a fine-tuning manner.
\subsection{Proxy Tasks} 
In a broad view of self-supervised learning, the proxy tasks can be constructed over multiple sensory data such as ego-motion and sound~\cite{SeeByMoving2015,AutoEncoding14,EgoMotion2017,LookListenLearn2016,LookListenLearn2017}. In a special view of visual representation learning, proxy tasks can be categorized into: (1) Image property transform and (2) Video content transform.

\subsubsection {Image Property Transform}
Spatial transforms applied on images can produce supervision signals for representation learning~\cite{larsson2017colorization}. As a representative research, \cite{gidaris2018unsupervised} proposed learning CNN features by rotating the images and predicting the rotated angles. \cite{kim2018learning,doersch2015unsupervised} proposed learning image representations by completing damaged Jigsaw puzzles. 
~\cite{Inpainting2016} proposed context inpainting, by training a CNN to generate the contents of a withheld image region conditioned on its surroundings.
~\cite{Invariance2017} proposed unsupervised correspondence, by training a representation model to match image patches of transform in-variance.

\subsubsection {Video Content Transform}
A large number of video clips with rich motion information provide various self-supervised signals. In ~\cite{WangXiaoLong2015}, the order of video frames was used as a supervisory signal. In ~\cite{misra2016shuffle,lee2017unsupervised}, predicting the orders of frames or video clips drives learning spatio-temporal representation. In~\cite{fernando2017self}, an odd-one-out network was proposed to identify the unrelated or odd video clips from a set of otherwise related clips. To find the odd video clip, the models have to learn spatio-temporal features that can discriminate similar video clips. In~\cite{WatchingMove2017}, unsupervised motion-based segmentation on videos was used to obtain segments, which performed as pseudo ground truth to train a CNN to segment objects.

Early works usually learned features based on 2D CNN and merely on a frame-by-frame basis, which are implausible to video analytic tasks where spatio-temporal features are prevailing.
Recently, ~\cite{wang2019self} proposed learning 3D representations by regressing motion and appearance statistics, ~\cite{xu2019self} proposed predicting the order of video clips. \cite{kim2019self} proposed training 3D CNN by completing space-time cubic puzzles.

However, existing self-supervised learning methods are typically designed for specific target tasks, which restricts the capability of learned models. In addition, few of the proxy tasks are capable of assessing feature representation, which strikingly limits the pertinence of learned models.

\begin{figure*}[!t]
     \centering
     \includegraphics[width=1.8\columnwidth]{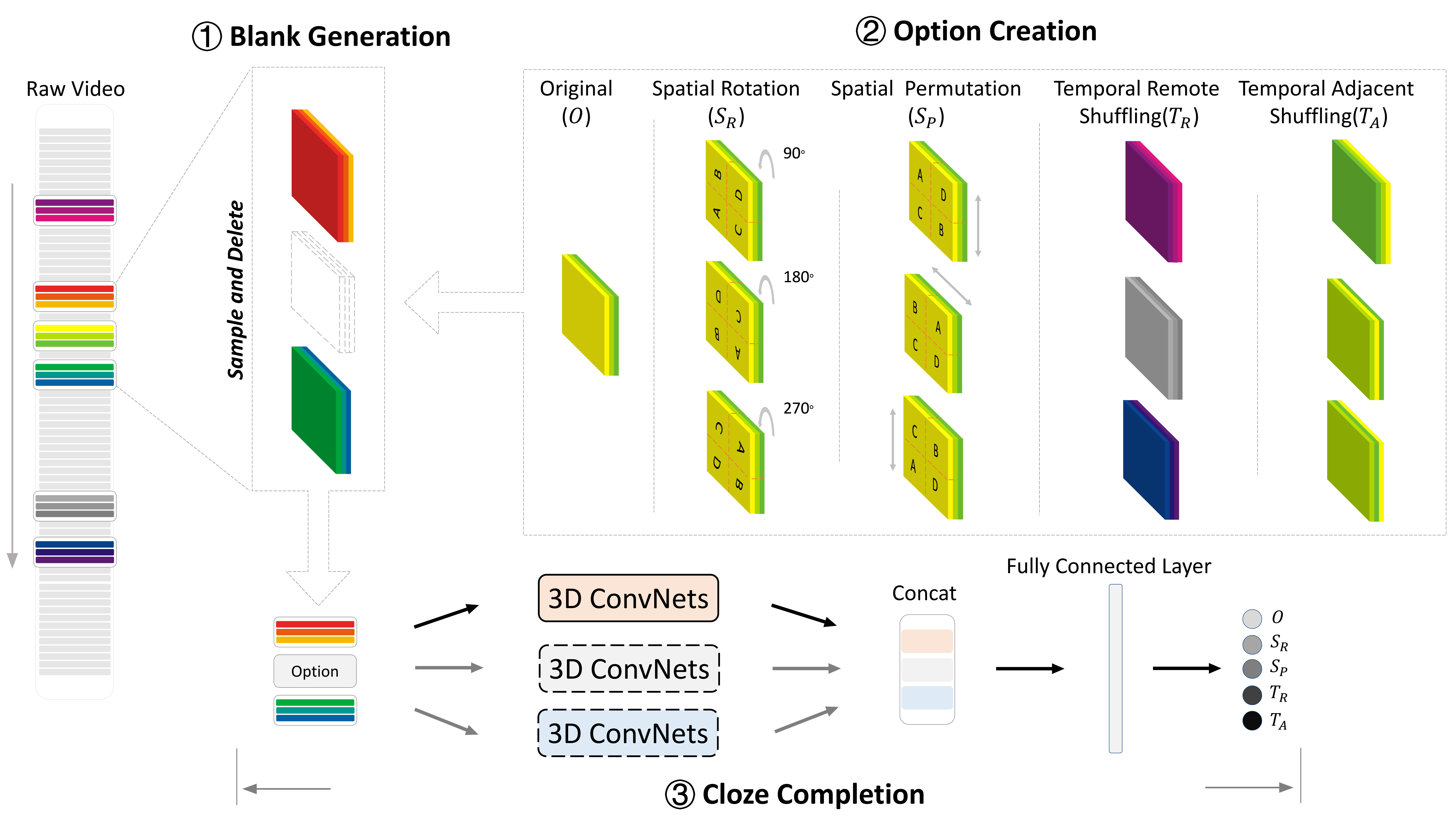}
     \caption{Illustration of the VCP framework. Given a video sequence, a sampled video clip is withheld and multiple spatio-temporal operations are applied on the withheld clip (up). A 3D-CNN model is applied to identify the category of operations, which drives learning rich feature representations. The motivation behinds VCP lies in that applying richer operations on the video clips facilities exploring richer feature representation (down). }
     \label{fig:framework}
 \end{figure*}

\subsection{Target Tasks} 
In this work, the self-supervised representation models are applied to target tasks including video action recognition and video retrieval. 
In many recent works, 
~\cite{tran2018closer,tran2015learning} investigated training 3D CNN models on a large scale supervised video database. 
Nevertheless, the models trained on specific self-supervised tasks lack general applicability, $i.e.$, fine-tuning such models to various video tasks could produce sub-optimal results. To conquer these issues, we propose the novel VCP, which, by incorporating multiple self-supervised representations, improves the generality of the learned model.

\section{Video Cloze Procedure}
Cloze Procedure was firstly introduced by Wilson Taylor in 1953 as a metric to evaluate the capability of human language learning. Specifically, it deletes words in a prose selection according to a word-count formula or various other criteria and evaluates the success a reader has in accurately supplying the deleted words \cite{bickley1970cloze}. Motivated by the success of Cloze Procedure in the field of language learning, we design the Video Cloze Procedure.

In this section, we first describe the details of VCP which consists of three components, $i.e.,$ blank generation, option creation, and cloze completion. We then discuss the advantages of the VCP over state-of-the-art methods in three aspects, including complexity, flexibility, and interpretability. 

\subsection{Blank Generation} 
Considering the spatial similarity and the temporal ambiguity among video frames, we take video clips~\cite{xu2019self} as the smallest unit in VCP. Considering that semantic information of different videos is temporally non-uniform, we generate the blanks in VCP using every-nth-words manner \cite{bickley1970cloze}. Specifically, the blank generation component consists of two steps including clip sampling and clip deletion.

\subsubsection{Clip Sampling} The clips including $k$ frames (with equal length) are sampled every $l$ frames (with equal interval without overlap) from the raw video. In this way, the relevance of the low-level vision cues, such as texture and color, among clips is weakened compared to those in successive or overlapped clips. As a result, the learner is forced to focus on middle- and high-level spatio-temporal features. 

\subsubsection{Clip Deletion} A video sequence of $m$ successive clips is considered as a whole cloze item. We randomly delete one of the co-equal clips with the same probability in the cloze item to generate blanks. The removed clip is then utilized to create options. For clarity of description, we give an example of VCP by sampling three clips and deleting the middle one, as shown in Fig.\ \ref{fig:framework}.

\subsection{Option Creation}
Aiming at training a model to distinguish the deleted clip from a heap of perplexing optional clips, we design spatial and temporal operations to create the optional clips (options). To learn richer representations, the operations should effectively confuse the learners, while reserving the spatial-temporal relevance. Under this principle, we design four operations including spatial rotation ($S_{R}$), spatial permutation ($S_{P}$), temporal remote shuffling ($T_{R}$), and temporal adjacent shuffling ($T_{A}$) for VCP. 

\subsubsection{Spatial Operation}
 To provide options that focus on spatial representation learning, we introduce spatial rotation and spatial permutation.
 With spatial rotation ($S_{R}$), a video clip is rotated by 90, 180, and 270 degrees so that the model is forced to learn orientation related features.
 With spatial permutation ($S_{P}$), a video clip is divided into four tiles ($2\times2\times1   grids)$ and either two tiles are permuted to produce a new option. There are $C_4^2=6$ kinds of options produced in total. Permutation with two tiles produces options with spatial structure information partially remained, which prevents models from learning low-level statistics to distinguish spatial chaos.
 
\subsubsection{Temporal Operation}
 To provide options that focus on temporal features we further introduce two kinds of temporal operations.
 One operation is temporal remote shuffling ($T_{R}$), where the deleted clip is substituted with a clip that has large temporal distance forward or backward. As the background of frames with reasonable temporal distance is probably similar which means the discriminative difference lies in the foreground, $T_{R}$ drives the model to learn more temporal information related to the foreground.
 The other operation is temporal adjacent shuffling ($T_{A}$), where the original clip is divided into four sub-clips, and two of them are randomly shuffled once. Different from VCOP \cite{xu2019self}, we do not shuffle all the sub-clips and reduce the difficulty by forcing the model to judge whether or not the clip is shuffled instead of predicting the exact orders. In this way, rich temporal representation can be easy to learn.

 \subsection{Cloze Completion}
To complete cloze, we fill the blanks by randomly sampling the clip options with operation category labels.
To predict the operation categories applied on the clips, we use three 3D CNNs as the backbones and concatenate their output features according to the order of the clips in the raw video as illustrated in  Fig.\ \ref{fig:framework}. 
The three CNNs share parameters so that a single strong model can be learned.
The concatenate feature is fed to a fully connected (FC) layer, which predicts the corresponding operation category. 

\begin{figure}[t]
     \centering
     \includegraphics[width=1\columnwidth]{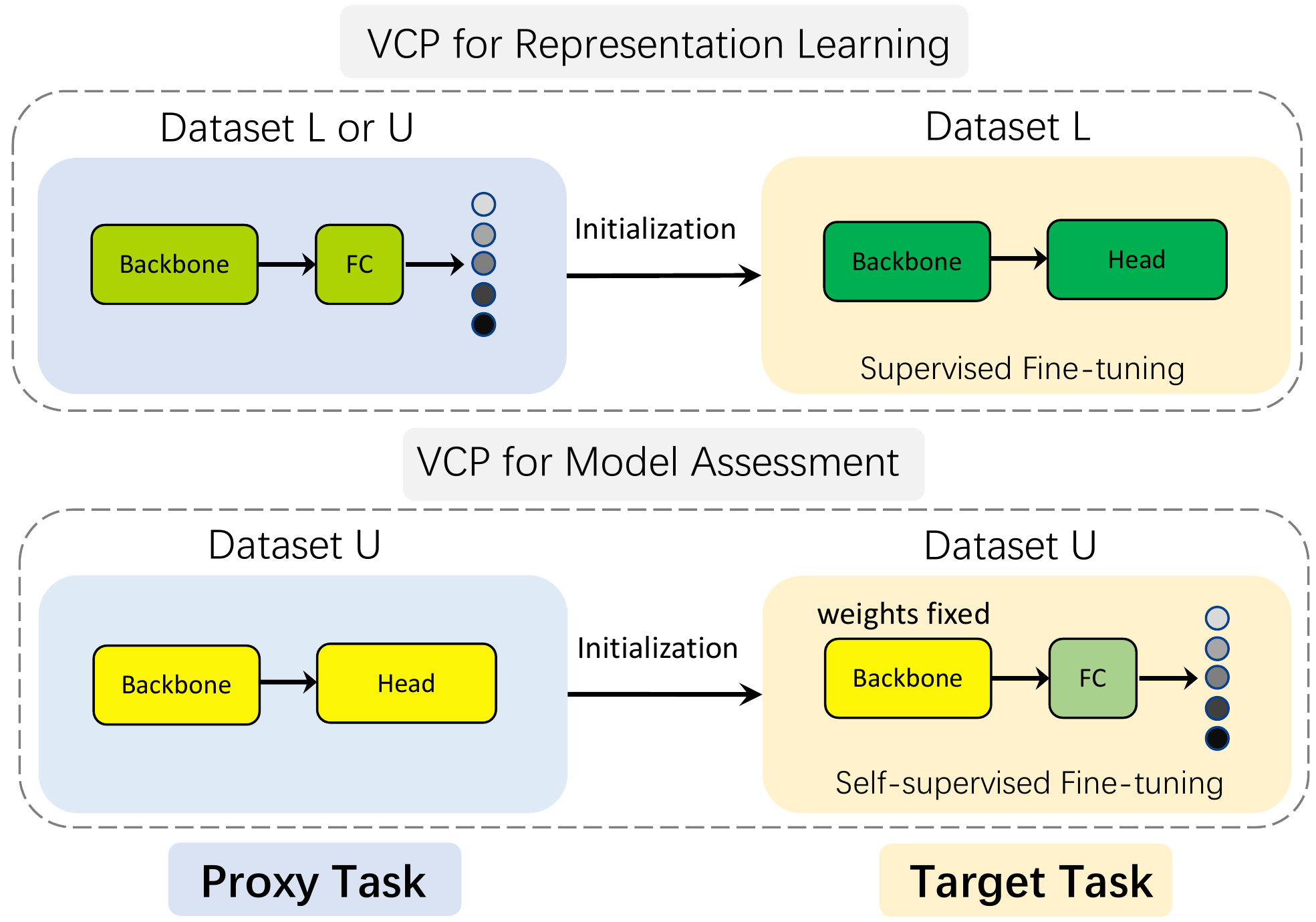}
     \caption{VCP can be utilized for representation learning with only the original labeled data (dataset L) for target tasks or using extra unlabeled data (dataset U). VCP can also act as a target task for model assessment, which can be used to evaluate self-supervised representation models. }
     \label{fig:VCPusage}
 \end{figure}
\section{Self-supervised Representation Learning}
We implement self-supervised representation learning and model assessment by treating VCP as a proxy task and a target task, respectively.

\subsection{Representation Learning}

As a proxy task, VCP can learn spatio-temporal representations using only the original labeled data for target tasks or using extra unlabeled data, Fig.\ \ref{fig:VCPusage}.

For the target task, deep models learn to extract features in a direct manner trying to minimize the training loss with the supervision of specific annotations, $i.e.,$ category labels. During the procedure, the task-specific representation capability of models can be enforced while the general representation capacity of models is unfortunately ignored. With spatio-temporal operations applied on the clips, VCP learns rich and general representations by pre-training the models, which enhances the performance of target tasks without extra labeling efforts required. 

On the other hand, VCP can leverage massive unlabeled data to break the overhead of model representation capability. With VCP, we pre-train a representation model on an un-annotated dataset as a warm-up initialization and then fine-tune such model on the annotated target dataset. VCP has the potential to learn the general representation, $e.g.,$ spatial-temporal integrity and continuity, in spatio-temporal domain, which facilitates improving the representation capability of models in video-based vision tasks. 

\subsection{Model Assessment} 
Beyond acting as a proxy task, VCP can also act as a target task, which offers a uniform and interpretable way to evaluate self-supervised representation models. In VCP, the classification accuracy of operations reflects what the models learn and how good they are. By simply replacing the head of the classification network with a fully connected layer to be fine-tuned while the parameters of the backbone network are fixed, operation category classification is implemented as a target task, Fig.\ \ref{fig:VCPusage}. 

In this way, the feature representative capability obtained from self-supervised proxy tasks is reserved. Meanwhile, corresponding features are utilized to train a classifier, the performance of which can be regarded as a metric to assess the representation models. With the hint dropped by VCP, we can not only elaborately assess models learned from different self-supervised proxy tasks but also can figure out how to improve a self-supervised method. This casts a new light on the significance of VCP.

\subsection{Discussion}
To analyze the advantages of VCP over existing self-supervised methods, we contrast them from three aspects including complexity, flexibility, and interpretability.

\subsubsection{Complexity} 
Existing approaches that use spatio-temporal shuffling and order prediction ~\cite{kim2019self,xu2019self,lee2017unsupervised} have $O(n!)$ computational complexity, given $n$ video frames/clips units. 
The high complexity is caused by the requirement to predict the exact order, which might be not necessary when learning representations.
In contrast, VCP solely chooses $n$ optional options to fill the blanks while predicting the operation category of the option. It thus has a $O(n)$ computational complexity. 

\subsubsection{Flexibility}
For various target tasks, VCP can be adaptively applied by configuring the options (operations). 
For example, we can apply spatial permutation ($S_{P}$) to enhance spatial representation and apply temporal adjacent shuffling ($T_{A}$) to boost the temporal representation. In a flexible manner, VCP can incorporate special information in special spatial and/or temporal operations for different target tasks.

\subsubsection{Interpretability} 
In existing approaches, different proxy tasks learn different representation models. It requires an interpretive way to explore the relationship between representation models and target tasks. 
With well-designed options, VCP offers the opportunity to analyze the models by testing their classification accuracy on uniform options (operations), which has great potential to contrapuntally overcome the weakness of models. 

\section{Experiments}
We conduct extensive experiments to evaluate VCP and its applications on target tasks.
Firstly, we elaborate experimental settings for VCP. We then evaluate the representation learning of VCP with different option configurations and data strategies. We further conduct experiments on model assessment with VCP. Finally, we evaluate the performance of VCP applying on target tasks, $i.e.$, action recognition and video retrieval, and compare it with state-of-the-art methods.

\subsection{Experiment Setting}
\subsubsection{Datasets} 
The experiments are conducted on UCF101 \cite{soomro2012ucf101} and HMDB51 \cite{jhuang2011large} datasets. UCF101 contains 13320 videos over 101 action categories, exhibiting challenging problems include intra-class variance of actions, complex camera motions, and cluttered backgrounds. HMDB51 contains 6849 videos over 51 action categories. The videos are mainly collected from movies and websites including the Prelinger archive, YouTube, and Google videos.

\subsubsection{Backbone Networks} 
 C3D \cite{tran2015learning}, R3D and R(2+1)D \cite{tran2018closer} are employed as backbones in VCP implementations. C3D extends the 2D convolution kernels to 3D kernels, so that it can model temporal information of videos. The size of convolution kernels 
is $3\times 3\times 3$. R3D is an extension of ResNet \cite{he2016deep} with C3D. In R(2+1)D, 3D convolution kernels are decomposed. For spatial convolution, each kernel is set to be $1\times n\times n$ where $n=3$. For temporal convolution, it is set to be $m\times 1\times 1$ where $m=3$.

\subsubsection{Implementation Details}
In the blank generation, to avoid trivial results, three successive 16-frame clips are sampled every 8 frames from the raw video as a whole cloze item. Each frame is resized to $128\times171$ and randomly cropped to $112\times112$. In the option generation, we define the clips sampled from 16 frames away to the cloze item as remote clips. We set the initial learning rate to be 0.01, momentum to be 0.9 and stop training after 300 epochs.

\begin{table}
    \resizebox{1.0\columnwidth}{!}{
    \centering
    \begin{tabular}{lcccccc}
    \hline
     3D CNNs&  Overall(\%)& $O$(\%) & $S_{R}$(\%) & $S_{P}$(\%) & $T_{R}$(\%)& $T_{A}$(\%)\\
    \hline
    C3D &78.42& 60.49& 95.04& 97.53  &47.32&94.57\\
    \hline
    \end{tabular}
    }
    \caption{Accuracy of operation classification. ``$O$'' denotes original video clips, ``$S_{R}$''  the spatial rotation, ``$S_{P}$'' the spatial permutation , ``$T_{R}$'' the temporal remote shuffling, and ``$T_{A}$'' the temporal adjacent shuffling .}
    \label{fig:vcp acc}
\end{table}

\begin{figure*}
     \centering
     \includegraphics[width=2.1\columnwidth]{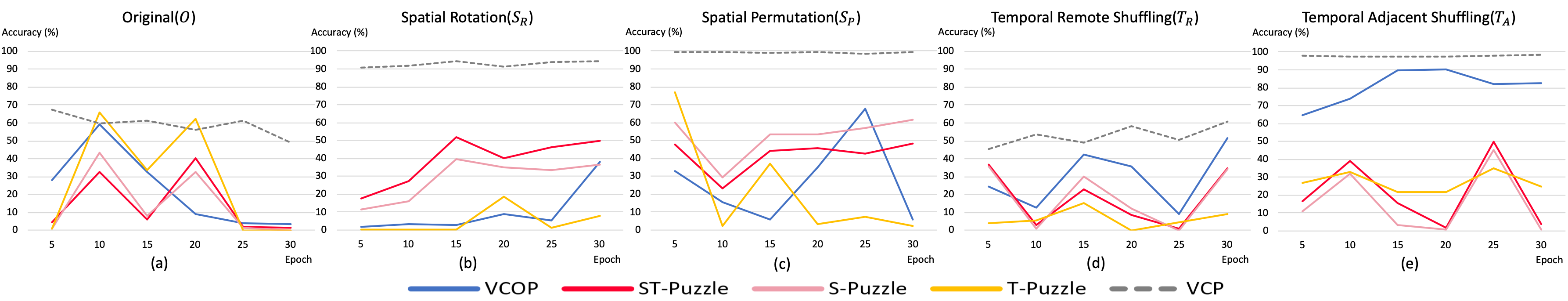}
     \caption{Model assessment results of VCOP \cite{xu2019self} and 3D Cubic puzzle \cite{kim2019self}. ``S-Puzzle" denotes spatial permutation, and  ``T-Puzzle" temporal permutation, ``ST-Puzzle" spatial and temporal permutation.}
     \label{fig:Video assessment}
 \end{figure*}
 
\subsection{Representation Learning}
To validate what VCP learns, we first conduct ablation studies of VCP. We further conduct experiments with different data strategies to demonstrate the generality of the representations learned via VCP.

\subsubsection{Ablation Study}
We firstly train a model to classify the categories of five options. Table \ref{fig:vcp acc} shows the results on UCF101, which are trained and  evaluated on the first split. It can be seen that VCP achieves 78.42\% overall accuracy, For spatial rotation ($S_R$), spatial permutation ($S_p$), and temporal adjacent shuffling ($T_A$), VCP respectively achieves 95.04\%, 97.53\% and 94.57\% accuracy. The results show that the designed five operations are plausible.

\begin{table}
    \centering
    \begin{tabular}{rcc}
        \hline
    Method  &UCF101(\%)  \\
    \hline
    random & 62.0 \\
        \hline
    $S_{R}$-VCP&64.3\\
    $S_{P}$-VCP&63.4\\
    $S_{R,P}$-VCP&66.0\\
        \hline
    $T_{R}$-VCP &67.8\\
    $T_{A}$-VCP & 65.0\\
    $T_{R,A}$-VCP&68.0\\
        \hline
   $ S_{R,P}T_{R,A}$-VCP&\textbf{69.7}\\
        \hline
        \end{tabular}
    \caption{Ablation study of spatio-temporal operations. The figures refer to action recognition accuracy.}
    \label{table:options}
\end{table}

To clearly show the effect of option creation for representation learning, we conduct ablation experiments on VCP with various options for action recognition, Table \ref{table:options}. The experiments are conducted using C3D as the backbone. We pre-train VCP and then fine-tune the action recognition model on UCF101. The recognition accuracy is evaluated on the first test split.

It can be seen that when pre-training with a single spatial rotation ($S_R$-VCP) or permutation ($S_P$-VCP) operation, the accuracy of action recognition outperforms the baseline (random) by  2.3\% or 1.4\%. When using both spatial operations ($S_{R,P}$-VCP), the performance further increased to 66.0\%. Pre-training with a single temporal remote shuffling ($T_R$-VCP) or adjacent shuffling ($T_A$-VCP) operation improves the performance by 5.8\% or 3.0\%, where the performance is further improved to 68.0\% when using both temporal operations ($T_{R,A}$-VCP). Combining the spatial and temporal operations ($S_{R,P}T_{R,A}$-VCP) finally improves the performance to 69.7\%, significantly outperforming the baseline by 7.7\%. The experiments show that the options can be used in a flexible way including using standalone or combining with each other. VCP can learn more representative features by adding rich and complementary options. 

\begin{table}
    \resizebox{1.0\columnwidth}{!}{
    \centering
    \begin{tabular}{lccc}
    \hline
    Data Strategy & VCOP~\cite{xu2019self} (\%) & VCP(ours) (\%)\\
    \hline
UCF101(random)&61.8&61.8\\
UCF101(UCF101)&65.6&\textbf{68.5}\\
UCF101(HMDB51)&64.1&\textbf{66.7}\\
    \hline
HMDB51(random)&24.7&24.7\\
HMDB51(HMDB51)&31.3&\textbf{31.5}\\
HMDB51(UCF101)&28.4&\textbf{32.5}\\
    \hline
        \end{tabular}
    }
    \caption{Performance (average on all test splits) comparison under different data strategies. UCF101 (HMDB51) dentoes the model is pre-trained on HMDB51 and fine-tuned on UCF101.}
    \label{table:cross_training}
\end{table}

\subsubsection{Data Strategy}
To further validate the generality of VCP, we conduct experiments for VCP under different data strategies, with C3D as the backbone. Firstly, we pre-train VCP on UCF101 and HMDB51, and then respectively fine-tune each pre-trained model on UCF101 and HMDB51 for action recognition, Table \ref{table:cross_training}. Specially, the supervised action recognition task is directly trained on the target datasets, with random initialization.

It can be seen that when pre-training and fine-tuning on UCF101, VCP outperforms VCOP by 2.9\%; when pre-training and fine-tuning on HMDB51, VCP slightly outperforms VCOP, showing that the strategy used in VCP is better than that in VCOP. Note that using VCP as a pre-train model further significantly improves the performance of supervised methods by 6.7\% (68.5\% vs. 61.8\%) on UCF101 and 6.8\% (31.5\% vs. 24.7\%) on HMDB51, which shows that VCP is complementary to supervised model learning. After pre-training on UCF101 and fine-tuning models on HMDB51, VCP significantly outperforms VCOP by 4.1\%. It is noteworthy that when pre-training on the smaller dataset HMDB51 but fine-tuning on the larger dataset UCF101, the performance of VCP also outperforms that of VCOP by 2.6\%, which shows the generality of VCP.

\begin{figure*}[t]
     \centering
     \includegraphics[width=2.0\columnwidth]{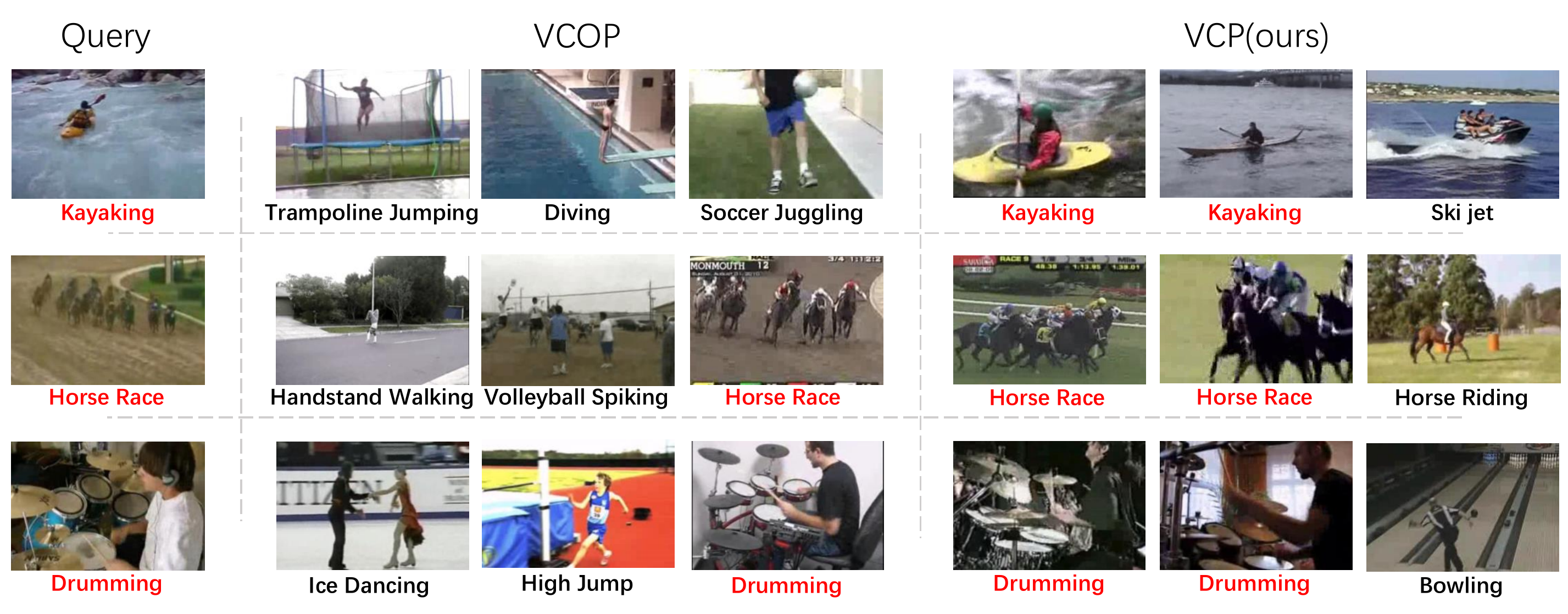}
     \caption{Comparison of video retrieval results. Red fonts indicate correct retrieval results. (Best viewed in color)}
     \label{fig:video retrieval}
 \end{figure*}
 
\subsection{Model Assessment}
Regarding VCP as a target task, we only fine-tune the fully connected layer with the parameters of the self-supervised model fixed to get the operation classification accuracy curve, Fig.\ \ref{fig:Video assessment}. We fine-tune the fully connected layer for 30 epochs and then output the test scores every 5 epochs.

It is obvious that the model trained with VCP can recognize the $S_{R}$, $S_{P}$, and $T_{A}$ operations with high accuracy ($\sim$90\%), Fig.\ \ref{fig:Video assessment}(b)(c)(e).
Nevertheless, it experiences difficulty when classifying the original clips and the remote shuffled clips, Fig.\ \ref{fig:Video assessment}(a)(d). 
It can be seen that the accuracy of $O$ and $T_{R}$ is negatively correlated, which means the perplexity of them.
In contrast, the accuracy of VCOP and 3D Cubic Puzzle is divergent, which implies they fail to classify the two categories.

For spatial operation classification, Fig.\ \ref{fig:Video assessment}(b)(c), ST-Puzzle and S-Puzzle outperform T-Puzze and VCOP, while for temporal operation classification, Fig.\ \ref{fig:Video assessment}(d)(e), they underperform T-Puzze and VCOP.
It shows that spatial representation learning is not consistent with temporal representation learning. Consequently, VCP benefits from integrating existing and newly designed spatial and temporal operations.

\subsection{Action Recognition}
\begin{table}
    \centering
        \resizebox{1.0\columnwidth}{!}{

    \begin{tabular}{lcc}
    \hline
    Method&UCF101(\%)&HMDB51(\%)  \\
    \hline
    Jigsaw\cite{noroozi2016unsupervised}&51.5&22.5 \\

    OPN\cite{lee2017unsupervised} &56.3&22.1\\
    B\"uchler\cite{buchler2018improving}  &58.6&25.0 \\
    Mas\cite{wang2019self} & 58.8&32.6\\
    3D ST-puzzle\cite{kim2019self}& 65.0 &31.3\\
    ImageNet pre-trained&67.1&28.5\\
    \hline
    C3D(random) &61.8&24.7\\
    C3D(VCOP\cite{xu2019self}) &65.6& 28.4\\ 
    C3D(VCP) &\textbf{68.5}&\textbf{32.5}\\ 
    \hline
    R3D(random) &54.5&23.4\\
    R3D(VCOP\cite{xu2019self}) &64.9& 29.5\\ 
    R3D(VCP) &\textbf{66.0}& \textbf{31.5}\\ 
    
    \hline
    R(2+1)D(random) &55.8&22.0\\
    R(2+1)D(VCOP\cite{xu2019self}) &72.4& 30.9\\ 
    R(2+1)D(VCP) &\textbf{66.3}& \textbf{32.2}\\ 
    \hline

    \end{tabular}
    }
    \caption{Comparison of action recognition accuracy on UCF101 and HMDB51.}
    \label{fig:state of the art}
\end{table}

Once a 3D CNN is pre-trained by VCP, we use it to initialize and fine-tune  models for action recognition.
For action recognition, we feed the features extracted by backbones to fully-connected layers for classification. During fine-tuning, we initialize the backbones from VCP while the fully-connected layers are randomly initialized. The hyper-parameters and data pre-processing are the same as VCP training process.
The fine-tune procedures are carried out for 150 epochs. During test, we follow the protocol of \cite{tran2018closer} and sample 10 clips for each video. The predictions on the clips are averaged to obtain the video prediction.

The classification accuracy over 3 splits are averaged to obtain the final accuracy.
As shown in Table \ref{fig:state of the art}, with a C3D backbone, VCP (ours) outperforms the randomly initialized C3D (random) by 6.7\% and 7.8\% on UCF101 and HMDB51 respectively. 
It also outperforms the state-of-the-art VCOP approach \cite{xu2019self} by 2.9\% and 4.1\%. 
With an R3D backbone, VCP has 11.5\% (54.5\% vs. 66\%) and 9.8\% (32.5\% vs. 27.4\%) performance gain over the random initialization (random) approach. It also outperforms the state-of-the-art VCOP \cite{xu2019self} with significant margins. The good performance validates that VCP can learn richer and more discriminative features than other methods.

\subsection{Video Retrieval}

VCP is also validated on the target task of nearest-neighbor video retrieval. As it does not require training data annotation, it largely relies on the pre-trained representation models. 
We conduct this experiment with the first split of UCF101, following the protocol in \cite{xu2019self}.
The model trained by VCP is to used to extract convolutional (conv5) features for all samples (videos) in the training and test sets. Each video in the test set is used to query  $k$ nearest videos from the training set. If a video of the same category is matched, a correct retrieval is counted.

It can be seen in Table \ref{fig:retrieve ucf101} and \ref{fig:retrieval hmdb} that 
VCP significantly outperforms the compared approaches on all evaluation metrics, $i.e.,$ top-1, top-5, top-10, top-20, and top-50 accuracy. In Fig.\ \ref{fig:video retrieval}, qualitative results also shows superiority of VCP.

\begin{table}
    \resizebox{1.0\columnwidth}{!}{
    \centering
    \begin{tabular}{lccccc}
    \hline
    Methods&top1(\%)&top5(\%)&top10(\%)&top20(\%)&top50(\%)  \\
    
        \hline
    Jigsaw\cite{noroozi2016unsupervised} &19.7&28.5&33.5&40.0&49.4 \\
    OPN\cite{lee2017unsupervised} &19.9&28.7&34.0&40.6&51.6 \\
    B\"uchler\cite{buchler2018improving} &25.7&36.2&42.2&49.2&59.5 \\
        \hline
    C3D(random) &16.7&27.5&33.7&41.4&53.0 \\
    
    C3D(VCOP\cite{xu2019self}) & 12.5 & 29.0&39.0&50.6&66.9\\

    C3D(VCP)&\textbf{17.3}&\textbf{31.5}&\textbf{42.0}&\textbf{52.6}&\textbf{67.7} \\
    \hline
    R3D(random) &9.9&18.9&26.0&35.5&51.9 \\
    R3D(VCOP\cite{xu2019self}) & 14.1 & 30.3&40.4&51.1&66.5\\

    R3D(VCP)&\textbf{ 18.6}&\textbf{33.6}&\textbf{42.5}&\textbf{53.5}&\textbf{68.1} \\
    \hline
    R(2+1)D(random) &10.6&20.7&27.4&37.4&53.1 \\
   R(2+1)D(VCOP\cite{xu2019self}) & 10.7 & 25.9&35.4&47.3&63.9\\

     R(2+1)D(VCP)& \textbf{19.9}&\textbf{33.7}&\textbf{42.0}&\textbf{50.5}&\textbf{64.4} \\
    \hline
    \end{tabular}
    }
    \caption{Video retrieval performance on UCF101.}
    \label{fig:retrieve ucf101}
\end{table}

\begin{table}
    \resizebox{1.0\columnwidth}{!}{
    \centering
    \begin{tabular}{lccccc}
    \hline
    Methods&top1(\%)&top5(\%)&top10(\%)&top20(\%)&top50(\%)  \\
        \hline
    C3D(random) &7.4&20.5&31.9&44.5&66.3 \\
    C3D(VCOP\cite{xu2019self}) & 7.4 & 22.6&34.4&48.5&70.1\\
    C3D(VCP) &\textbf{7.8}&\textbf{23.8}&\textbf{35.3}&\textbf{49.3}&\textbf{71.6}\\

    \hline
    R3D(random) &6.7&18.3&28.3&43.1&67.9 \\
    R3D(VCOP\cite{xu2019self}) & 7.6 & 22.9&34.4&48.8&68.9\\
    R3D(VCP) & \textbf{7.6}&\textbf{24.4}&\textbf{36.3}&\textbf{53.6}&\textbf{76.4} \\

    \hline
            
    R(2+1)D(random) &4.5&14.8&23.4&38.9&63.0 \\
    R(2+1)D(VCOP\cite{xu2019self}) & 5.7 & 19.5&30.7&45.8&67.0\\

    R(2+1)D(VCP)&\textbf{6.7}&\textbf{21.3}&\textbf{32.7}&\textbf{49.2}&\textbf{73.3} \\
    \hline
    \end{tabular}
    }
    \caption{Video retrieval performance on HMDB51.}
    \label{fig:retrieval hmdb}
\end{table}
 
\section{Conclusion}
In this paper, we propose a novel self-supervised method, referred to as Video Cloze Procedure (VCP), to learn rich spatial-temporal representations. 
 With VCP, we train spatial-temporal representation models (3D-CNNs) and apply such models on action recognition and video retrieval tasks. 
 We also proposed a model assessment approach by designing VCP as a special target task, which improves the pertinence of self-supervised representation learning. 
 Experimental results validated that VCP enhanced the representation capability and the interpretability of self-supervised models. The underlying fact is that VCP simulates the fashion of human language learning, which provides a fresh insight for self-supervised learning tasks.

\section*{Acknowledgment}
This work is supported by the National Key R\&D Program of China
(2017YFB1002400)
and the Strategic Priority Research Program of Chinese Academy of Sciences
(XDC02000000)

\bibliographystyle{aaai}
\bibliography{2527-References}

\begin{thebibliography}{}

\bibitem[\protect\citeauthoryear{Bickley, Ellington, and
  Bickley}{1970}]{bickley1970cloze}
Bickley, A.; Ellington, B.~J.; and Bickley, R.~T.
\newblock 1970.
\newblock The cloze procedure: A conspectus.
\newblock {\em Journal of Reading Behavior} 2(3):232--249.

\bibitem[\protect\citeauthoryear{Buchler, Brattoli, and
  Ommer}{2018}]{buchler2018improving}
Buchler, U.; Brattoli, B.; and Ommer, B.
\newblock 2018.
\newblock Improving spatiotemporal self-supervision by deep reinforcement
  learning.
\newblock In {\em Proceedings of the European Conference on Computer Vision},
  770--786.

\bibitem[\protect\citeauthoryear{Deepak \bgroup et al\mbox.\egroup
  }{2016}]{Inpainting2016}
Deepak, P.; Philipp, K.; Jeff, D.; Trevor, D.; and Alexei, A.~E.
\newblock 2016.
\newblock Context encoders: Feature learning by inpainting.
\newblock In {\em Proceedings of the IEEE International Conference on Computer
  Vision and Pattern Recognition},  2536--2544.

\bibitem[\protect\citeauthoryear{Deepak \bgroup et al\mbox.\egroup
  }{2017}]{WatchingMove2017}
Deepak, P.; Ross, B.~G.; Piotr, D.; Trevor, D.; and Bharath, H.
\newblock 2017.
\newblock Learning features by watching objects move.
\newblock In {\em Proceedings of the IEEE International Conference on Computer
  Vision and Pattern Recognition},  6024--6033.

\bibitem[\protect\citeauthoryear{Diederik and Max}{2014}]{AutoEncoding14}
Diederik, P.~K., and Max, W.
\newblock 2014.
\newblock Auto-encoding variational bayes.
\newblock In {\em Proceedings of International Conference on Learning
  Representations}.

\bibitem[\protect\citeauthoryear{Dinesh and Kristen}{2017}]{EgoMotion2017}
Dinesh, J., and Kristen, G.
\newblock 2017.
\newblock Learning image representations tied to egomotion from unlabeled
  video.
\newblock {\em International Journal of Computer Vision} 125(1-3):136--161.

\bibitem[\protect\citeauthoryear{Doersch, Gupta, and
  Efros}{2015}]{doersch2015unsupervised}
Doersch, C.; Gupta, A.; and Efros, A.~A.
\newblock 2015.
\newblock Unsupervised visual representation learning by context prediction.
\newblock In {\em Proceedings of the IEEE International Conference on Computer
  Vision},  1422--1430.

\bibitem[\protect\citeauthoryear{Fernando \bgroup et al\mbox.\egroup
  }{2017}]{fernando2017self}
Fernando, B.; Bilen, H.; Gavves, E.; and Gould, S.
\newblock 2017.
\newblock Self-supervised video representation learning with odd-one-out
  networks.
\newblock In {\em Proceedings of the IEEE Conference on Computer Vision and
  Pattern Recognition},  3636--3645.

\bibitem[\protect\citeauthoryear{Gidaris, Singh, and
  Komodakis}{2018}]{gidaris2018unsupervised}
Gidaris, S.; Singh, P.; and Komodakis, N.
\newblock 2018.
\newblock Unsupervised representation learning by predicting image rotations.
\newblock {\em arXiv preprint arXiv:1803.07728}.

\bibitem[\protect\citeauthoryear{He \bgroup et al\mbox.\egroup
  }{2016}]{he2016deep}
He, K.; Zhang, X.; Ren, S.; and Sun, J.
\newblock 2016.
\newblock Deep residual learning for image recognition.
\newblock In {\em Proceedings of the IEEE Conference on Computer Vision and
  Pattern Recognition},  770--778.

\bibitem[\protect\citeauthoryear{Jhuang \bgroup et al\mbox.\egroup
  }{2011}]{jhuang2011large}
Jhuang, H.; Garrote, H.; Poggio, E.; Serre, T.; and Hmdb, T.
\newblock 2011.
\newblock A large video database for human motion recognition.
\newblock In {\em Proceedings of the IEEE International Conference on Computer
  Vision}, volume~4, ~6.

\bibitem[\protect\citeauthoryear{Jia \bgroup et al\mbox.\egroup
  }{2009}]{ImageNet16}
Jia, D.; Wei, D.; Richard, S.; Li{-}Jia, L.; Kai, L.; and Fei{-}Fei, L.
\newblock 2009.
\newblock Imagenet: {A} large-scale hierarchical image database.
\newblock In {\em Proceedings of the IEEE International Conference on Computer
  Vision and Pattern Recognition},  248--255.

\bibitem[\protect\citeauthoryear{Jimmy \bgroup et al\mbox.\egroup
  }{2016}]{LookListenLearn2016}
Jimmy, S. J.~R.; Yongtao, H.; Yu{-}Wing, T.; Chuan, W.; Li, X.; Wenxiu, S.; and
  Qiong, Y.
\newblock 2016.
\newblock Look, listen and learn- {A} multimodal {LSTM} for speaker
  identification.
\newblock In {\em Proceedings of the Thirtieth Conference on Artificial
  Intelligence},  3581--3587.

\bibitem[\protect\citeauthoryear{Kay \bgroup et al\mbox.\egroup
  }{2017}]{kay2017kinetics}
Kay, W.; Carreira, J.; Simonyan, K.; Zhang, B.; Hillier, C.; Vijayanarasimhan,
  S.; Viola, F.; Green, T.; Back, T.; Natsev, P.; et~al.
\newblock 2017.
\newblock The kinetics human action video dataset.
\newblock {\em arXiv preprint arXiv:1705.06950}.

\bibitem[\protect\citeauthoryear{Kim \bgroup et al\mbox.\egroup
  }{2018}]{kim2018learning}
Kim, D.; Cho, D.; Yoo, D.; and Kweon, I.~S.
\newblock 2018.
\newblock Learning image representations by completing damaged jigsaw puzzles.
\newblock In {\em 2018 IEEE Winter Conference on Applications of Computer
  Vision (WACV)},  793--802.
\newblock IEEE.

\bibitem[\protect\citeauthoryear{Kim, Cho, and Kweon}{2019}]{kim2019self}
Kim, D.; Cho, D.; and Kweon, I.~S.
\newblock 2019.
\newblock Self-supervised video representation learning with space-time cubic
  puzzles.
\newblock In {\em Proceedings of the AAAI Conference on Artificial
  Intelligence}, volume~33,  8545--8552.

\bibitem[\protect\citeauthoryear{Larsson, Maire, and
  Shakhnarovich}{2017}]{larsson2017colorization}
Larsson, G.; Maire, M.; and Shakhnarovich, G.
\newblock 2017.
\newblock Colorization as a proxy task for visual understanding.
\newblock In {\em Proceedings of the IEEE Conference on Computer Vision and
  Pattern Recognition},  6874--6883.

\bibitem[\protect\citeauthoryear{Lee \bgroup et al\mbox.\egroup
  }{2017}]{lee2017unsupervised}
Lee, H.-Y.; Huang, J.-B.; Singh, M.; and Yang, M.-H.
\newblock 2017.
\newblock Unsupervised representation learning by sorting sequences.
\newblock In {\em Proceedings of the IEEE International Conference on Computer
  Vision},  667--676.

\bibitem[\protect\citeauthoryear{Misra, Zitnick, and
  Hebert}{2016}]{misra2016shuffle}
Misra, I.; Zitnick, C.~L.; and Hebert, M.
\newblock 2016.
\newblock Shuffle and learn: unsupervised learning using temporal order
  verification.
\newblock In {\em European Conference on Computer Vision},  527--544.
\newblock Springer.

\bibitem[\protect\citeauthoryear{Noroozi and
  Favaro}{2016}]{noroozi2016unsupervised}
Noroozi, M., and Favaro, P.
\newblock 2016.
\newblock Unsupervised learning of visual representations by solving jigsaw
  puzzles.
\newblock In {\em European Conference on Computer Vision},  69--84.
\newblock Springer.

\bibitem[\protect\citeauthoryear{Pulkit, Jo{\~{a}}o, and
  Jitendra}{2015}]{SeeByMoving2015}
Pulkit, A.; Jo{\~{a}}o, C.; and Jitendra, M.
\newblock 2015.
\newblock Learning to see by moving.
\newblock In {\em Proceedings of the IEEE International Conference on Computer
  Vision},  37--45.

\bibitem[\protect\citeauthoryear{Relja and Andrew}{2017}]{LookListenLearn2017}
Relja, A., and Andrew, Z.
\newblock 2017.
\newblock Look, listen and learn.
\newblock In {\em Proceedings of the IEEE International Conference on Computer
  Vision},  609--617.

\bibitem[\protect\citeauthoryear{Soomro, Zamir, and
  Shah}{2012}]{soomro2012ucf101}
Soomro, K.; Zamir, A.~R.; and Shah, M.
\newblock 2012.
\newblock Ucf101: A dataset of 101 human actions classes from videos in the
  wild.
\newblock {\em arXiv preprint arXiv:1212.0402}.

\bibitem[\protect\citeauthoryear{Tran \bgroup et al\mbox.\egroup
  }{2015}]{tran2015learning}
Tran, D.; Bourdev, L.; Fergus, R.; Torresani, L.; and Paluri, M.
\newblock 2015.
\newblock Learning spatiotemporal features with 3d convolutional networks.
\newblock In {\em Proceedings of the IEEE International Conference on Computer
  Vision},  4489--4497.

\bibitem[\protect\citeauthoryear{Tran \bgroup et al\mbox.\egroup
  }{2018}]{tran2018closer}
Tran, D.; Wang, H.; Torresani, L.; Ray, J.; LeCun, Y.; and Paluri, M.
\newblock 2018.
\newblock A closer look at spatiotemporal convolutions for action recognition.
\newblock In {\em Proceedings of the IEEE Conference on Computer Vision and
  Pattern Recognition},  6450--6459.

\bibitem[\protect\citeauthoryear{Wang \bgroup et al\mbox.\egroup
  }{2019}]{wang2019self}
Wang, J.; Jiao, J.; Bao, L.; He, S.; Liu, Y.; and Liu, W.
\newblock 2019.
\newblock Self-supervised spatio-temporal representation learning for videos by
  predicting motion and appearance statistics.
\newblock In {\em Proceedings of the IEEE Conference on Computer Vision and
  Pattern Recognition},  4006--4015.

\bibitem[\protect\citeauthoryear{Xiaolong and Abhinav}{2015}]{WangXiaoLong2015}
Xiaolong, W., and Abhinav, G.
\newblock 2015.
\newblock Unsupervised learning of visual representations using videos.
\newblock In {\em Proceedings of the IEEE International Conference on Computer
  Vision and Pattern Recognition},  2794--2802.

\bibitem[\protect\citeauthoryear{Xiaolong, Kaiming, and
  Abhinav}{2017}]{Invariance2017}
Xiaolong, W.; Kaiming, H.; and Abhinav, G.
\newblock 2017.
\newblock Transitive invariance for self-supervised visual representation
  learning.
\newblock In {\em Proceedings of the IEEE International Conference on Computer
  Vision and Pattern Recognition},  1338--1347.

\bibitem[\protect\citeauthoryear{Xu \bgroup et al\mbox.\egroup
  }{2019}]{xu2019self}
Xu, D.; Xiao, J.; Zhao, Z.; Shao, J.; Xie, D.; and Zhuang, Y.
\newblock 2019.
\newblock Self-supervised spatiotemporal learning via video clip order
  prediction.
\newblock In {\em Proceedings of the IEEE Conference on Computer Vision and
  Pattern Recognition},  10334--10343.

\end{thebibliography}

\end{document}